\documentclass[11pt]{article}
\usepackage[defaultlines=3,all]{nowidow}
\usepackage{graphicx} 
\usepackage{amsmath}
\usepackage{amsfonts}

\usepackage{geometry}
\usepackage{natbib}

\newcommand\OURMETHOD{\texttt{SemanticRegen}}
\usepackage{xcolor}
\usepackage{booktabs}
\usepackage{multirow}

\usepackage{hyperref}
\hypersetup{
    colorlinks,
    linkcolor={red!50!black},
    citecolor={magenta},
    urlcolor={blue!80!black}
}

\usepackage{geometry}
\title{Removing Watermarks with Partial Regeneration \\ using Semantic Information} 
\usepackage{authblk}

\everypar{\looseness=-1}

\author[1]{Krti Tallam}
\author[2]{John Kevin Cava}
\author[3]{Caleb Geniesse}
\author[1,3]{N. Benjamin Erichson}
\author[1,3,4]{\\Michael W. Mahoney}
\affil[1]{International Computer Science Institute, Berkeley, CA, USA}
\affil[2]{School of Computing and Augmented Intelligence, Arizona State University, AZ, USA}
\affil[3]{Lawrence Berkeley National Laboratory, Berkeley, CA, USA}
\affil[4]{Department of Statistics, University of California at Berkeley, Berkeley, CA, USA}

\date{}

\usepackage{parskip}

\begin{document}
\sloppy

\maketitle

\begin{abstract}
As AI-generated imagery becomes ubiquitous, invisible watermarks have emerged as a primary line of defense for copyright and provenance.  The newest watermarking schemes embed \emph{semantic} signals - content-aware patterns that are designed to survive common image manipulations - yet their true robustness against adaptive adversaries remains under-explored. We expose a previously unreported vulnerability and introduce \texttt{SemanticRegen}, a three-stage, label-free attack that erases state-of-the-art semantic and invisible watermarks while leaving an image’s apparent meaning intact.  Our pipeline (i) uses a vision-language model to obtain fine-grained captions, (ii) extracts foreground masks with zero-shot segmentation, and (iii) inpaints only the background via an LLM-guided diffusion model, thereby preserving salient objects and style cues. Evaluated on \textgreater{}1,000 prompts across four watermarking systems - TreeRing, StegaStamp, StableSig, and DWT/DCT - \texttt{SemanticRegen} is the \emph{only} method to defeat the semantic TreeRing watermark ($p\!=\!0.10{>}0.05$) and reduces bit-accuracy below 0.75 for the remaining schemes, all while maintaining high perceptual quality (masked SSIM = 0.94 ± 0.01).  We further introduce \emph{masked SSIM} (mSSIM) to quantify fidelity within foreground regions, showing that our attack achieves up to 12 percent higher mSSIM than prior diffusion-based attackers. These results highlight an urgent gap between current watermark defenses and the capabilities of adaptive, semantics-aware adversaries, underscoring the need for watermarking algorithms that are resilient to content-preserving regenerative attacks.
\end{abstract}
\vspace{+1cm}
\section{Introduction}
\label{sec:intro}
\linepenalty=1000

The growing advancement and widespread adoption of AI-generated content has brought about urgent challenges in protecting copyright and intellectual property, particularly in fields such as science, healthcare, and entertainment \cite{bohr2020rise,soori2023artificial}. As the reliance on these generated images grows, so does the need for robust methods to ensure the integrity and ownership of digital content \cite{gaur2024extensive,arrigo2023quantitative}. Watermarking embeds markers in images to verify ownership and prevent misuse \cite{fernandez2023stable,wen2023treering,2019stegastamp,cox2007digital,al2007combined,zhang2019robust}. Traditional watermarking techniques have been instrumental in the embedding of markers to verify ownership and prevent misuse \cite{sharma2023review,kumar2023compreshensive}. However, the rise of sophisticated adversarial attacks that subtly alter images to evade detection has exposed vulnerabilities in these systems \cite{saberi2024robustness,lyu2023adversarial,qiao2023novel,aberna2024digital}. This growing threat highlights the critical need for more advanced and resilient approaches to safeguarding digital assets \cite{qi2024investigating,chen2024deep}, ensuring that the rights of content creators are upheld in the face of evolving technological challenges.

\begin{figure}[!t]
\centering
\includegraphics[width=0.8\textwidth]{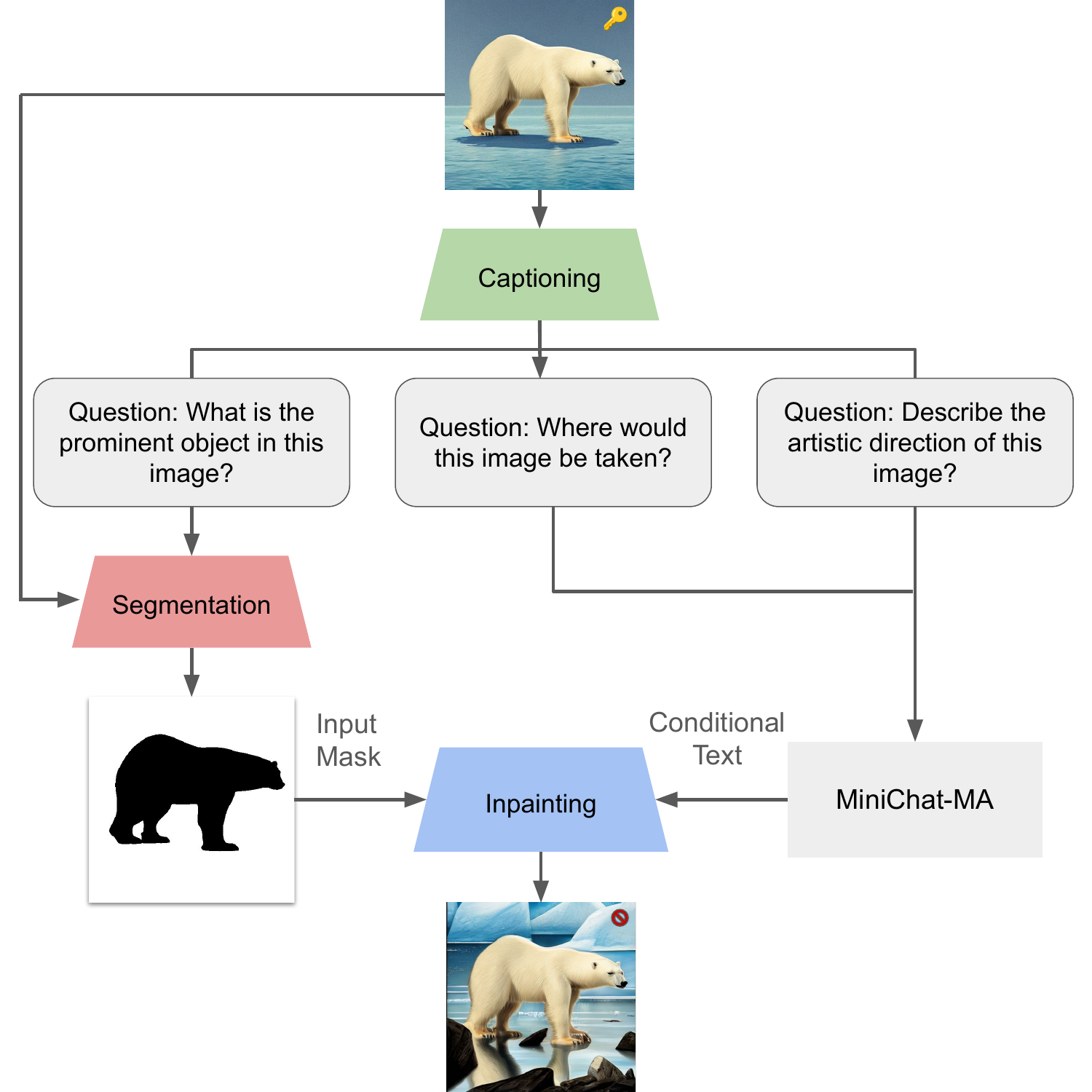}
\caption{Our semantic watermark removal pipeline involves three primary components: (1) Captioning (green), (2) Segmentation (red), and (3) Inpainting (blue). For captioning, we use a VQA model to provide essential context for subsequent processing. For segmentation, we focus on prominent objects or areas of interest within the image. For inpainting, the background of the image is replaced with semantically similar content, effectively removing the watermark while preserving image integrity. To construct the prompt for conditional text inpainting, we use MiniChat-MA, an LLM that refines answers generated from the image captioning model. 
This pipeline extracts semantic information and replaces the background for watermark removal, while preserving the foreground content.}
\label{fig:fig2}
\end{figure}

To address these challenges, researchers have developed techniques to embed markers in generated images to verify ownership and prevent unauthorized use \cite{begum2020digital,kannan2015extensive,tao2014robust}.
One such method involves using variable autoencoders (VAE) or diffusion models to inject watermarks into the latent space by encoding neural networks or adding noise, as demonstrated in the WAVES benchmark \cite{an2024benchmarking}. 
Zhao et al. proposed a watermark attack method, highlighting the need to strengthen watermarking strategies \cite{zhao2023invisible}. 
Adversaries have created advanced attacks to bypass detection by modifying subtly watermarked images, leading to the development of more robust detection systems \cite{ballé2018variational,cheng2020learned,saberi2024robustness,zhao2023invisible,an2024benchmarking}.
Advances in AI security have led to innovative strategies to protect generated images from adversarial manipulation \cite{kumar2023artificial}. Classifier-free methods for the detection and removal of watermarks offer alternatives to traditional approaches by analyzing the inherent properties of the image rather than relying on predefined classifiers \cite{zhong2023brief}. Techniques such as pixel-level analysis, frequency domain analysis, and structural analysis identify anomalies introduced by watermarks \cite{zhong2023brief,nadimpalli2023proactive}.

In this work, we propose \OURMETHOD, a framework for removing semantic watermarks; see Figure~\ref{fig:fig2} for an illustration of our basic approach. 
Our approach demonstrates the effectiveness of semantic repainting for watermark removal, exposing vulnerabilities that can inform the development of more resilient techniques. 
Our approach involves a three-step pipeline: (1) a Visual Question Answering (VQA) model through BLIP2; (2) a segmentation model using LangSAM; and (3) Stable Diffusion Inpainting \cite{li2023blip2,langsam,nie2022diffusion} (see Figure \ref{fig:fig2}). 
The VQA model analyzes the target image, providing semantic information.
Our method preserves the semantic background information instead of substituting it with random/arbitrary content. 
After generating prompts, we apply inverted masks from the segmentation model to condition the target image, creating a new image that retains salient objects while replacing the background with semantically similar content from the original image.
We use a comprehensive approach with the VQA model for content extraction, employing customized questions to capture diverse aspects of complex scenes. 
Conditioning the model with prompts improves its ability to discern details, helping to remove strong watermarks while preserving the integrity of the image \cite{zhong2023brief}. Unlike prior methods that rely solely on adversarial or generative transformations, our approach integrates semantic understanding via VQA-driven segmentation to improve targeted watermark removal while preserving content integrity.

Our approach is inspired by research on the use of large language models (LLMs) for synthetic data set generation and image diffusion models for robust training \cite{dunlap2023diversify,kawar2023imagic}. 
Recent advances in diffusion-based watermarking by Zhang et al. \cite{zhang2024robust} and Kawar et al. \cite{kawar2023imagic} show promise in embedding watermarks into images, while preserving visual fidelity. 
The WAVES benchmark \cite{an2024benchmarking} provides insights into their performance. Diffusion-based approaches prioritize image fidelity through controlled noise application, making them effective for watermark removal and preferable to GAN-based methods due to their stability, robustness, and ease of implementation. 
These techniques offer a compelling solution for various applications.
We evaluated \OURMETHOD against these and related methods, in particular against TreeRing watermarker \cite{wen2023treering}, StegaStamp \cite{2019stegastamp}, StableSig \cite{fernandez2023stable} and invisible watermarkers \cite{zhao2023invisible}. TreeRing is ideal due to its imperceptibility and resilience to common manipulations such as cropping, resizing, and compression \cite{wen2023treering}.

Our evaluation in various watermarking techniques demonstrates the effectiveness of \OURMETHOD, with minimal distortion and high image quality, as reflected in low Mean Squared Error (MSE) and high Structural Similarity Index (SSIM) and Peak Signal-to-Noise Ratio (PSNR) scores. 
Competing well against state-of-the-art techniques such as DiffWMAttacker, VAEWMAttacker and Rinse4x, \OURMETHOD fills removed portions with semantically similar backgrounds, achieving meaningful results. 
Optimal performance is achieved with clear, separable backgrounds similar to those in Stable Diffusion training data.
\OURMETHOD\ excels in watermark removal when the central meaning of the image is localized to only a few main objects, since the method relies on segmenting these objects and removing the less important background content.

\vspace{8pt}\noindent
In summary, our main contributions are as follows. 

\begin{itemize}
\item 
We introduce \OURMETHOD, an effective watermark removal method, evaluate it on the TreeRing watermarker \cite{wen2023treering}, and compare it to StegaStamp \cite{2019stegastamp}, StableSig \cite{fernandez2023stable}, and invisible watermarks of deep neural networks \cite{zhao2023invisible}. 
\OURMETHOD\ successfully removes all four watermarks tested, as demonstrated by our analyzes ($p$ $>$ $0.05$ and $Bit Accuracy$ $<$ $0.75$).
\item 
We compare \OURMETHOD with other watermark attackers and observe that \OURMETHOD is the only attacker to eliminate the semantic Tree-Ring watermark. 
\OURMETHOD\ has an average $p$-value of $0.1$, while all other watermark attackers failed to meet the threshold for successful removal ($p > 0.05$).

\item 
We introduce a metric to evaluate objects in the foreground: the masked structural similarity index (mSSIM). 
\OURMETHOD\ significantly outperforms current baseline methods in different watermarking methods, including invisible and semantic watermarks. 
Our method preserves image quality with the most success within salient regions of generated images, as evidenced by mSSIM scores of 0.94, compared to Image Distortion ($mSSIM = 0.85$) and Rinse4x ($mSSIM = 0.86$).
\item 
We demonstrate how SemanticRegen leverages a multi-step pipeline to expose and exploit vulnerabilities in current watermarking techniques. 
Using a VQA model for context generation, segmenting key areas of the image and replacing the background with semantically similar content through LLM-guided inpainting, \OURMETHOD extracts and reconstructs portions of the watermarked image. 
This process reveals latent patterns that these watermarking techniques fail to protect, offering insights into potential attack vectors and relevant underlying assumptions that could be exploited.\footnote{https://github.com/KrtiT/semanticRegen}
\end{itemize}

\section{Related Work}

Watermarking and its adversarial counterpart, watermark removal, are pivotal areas of research in protecting intellectual property and ensuring the integrity of AI-generated content. This section discusses advances in watermarking methods, challenges posed by adversarial attacks, and the broader implications for AI-generated media.

\paragraph{Watermarking Methods.}
Watermarking has evolved from traditional techniques, such as frequency domain embedding, to state-of-the-art methods that take advantage of generative models for imperceptible yet robust integration. Early approaches embedded watermarks in spatial or frequency domains using the discrete wavelet transform (DWT) or the discrete cosine transform (DCT) \cite{cox2007digital, al2007combined}. Although effective for basic transformations, these methods were vulnerable to adversarial attacks that exploited predictable patterns.

Modern watermarking techniques, such as TreeRing \cite{wen2023treering} and StegaStamp \cite{2019stegastamp}, have introduced imperceptible and resilient watermarks. 
TreeRing watermarks use adaptive encoding mechanisms to maintain the integrity of the watermark against manipulations such as resizing, cropping, and compression. 
Similarly, StegaStamp employs neural networks to embed and extract watermarks with high fidelity, enabling robust ownership verification.

Recent innovations leverage diffusion-based models to embed watermarks in image data while preserving visual fidelity \cite{zhang2024robust, kawar2023imagic}. These approaches integrate watermarks directly into the latent spaces of generative models, ensuring resilience against adversarial manipulations. In particular, \cite{arxiv2304.06790} explores how imperceptible signatures can be embedded in high-resolution generative content, paving the way for secure watermarking in multimodal AI systems.

Efforts such as the WAVES benchmark \citep{an2024benchmarking} have standardized the evaluation of watermarking techniques. WAVES provides a baseline framework to assess robustness across various attacks, offering insights into strengths and limitations. Such benchmarks are instrumental in the development of next-generation watermarking systems.

\paragraph{Watermark Removal Methods.}
The increasing sophistication of adversarial techniques has highlighted vulnerabilities in watermarking systems. Watermark removal methods exploit the inherent structure of embedded watermarks to obscure, distort, or eliminate them. Early approaches relied on pixel-level transformations, but recent advances employ machine learning techniques to target latent representations of watermarked content.

Regenerative attacks, such as those using Variational Autoencoders (VAEs) \cite{ballé2018variational, cheng2020learned} and diffusion models \cite{zhao2023invisible}, have proven effective in bypassing watermarking schemes. These methods iteratively refine watermarked images, reconstructing their features while removing embedded signals. 
In particular, Zhao et al. \cite{zhao2023invisible} demonstrate how diffusion-based methods can obscure watermarks while maintaining image fidelity, highlighting the need for continual innovation in watermarking strategies.

Hybrid approaches have also emerged that combine adversarial purification with iterative refinement techniques such as ``rinsing'' \cite{an2024benchmarking}. These methods sequentially reduce watermark detectability by applying regenerative transformations. For example, hybrid methods leverage both semantic understanding and low-level noise removal to effectively erase watermarks without compromising image quality \cite{nie2022diffusion, desu2024generative}. Furthermore, \cite{arxiv2304.06790} explores adversarial frameworks specifically designed to manipulate the robustness of the watermark, while \cite{arxiv2411.18479} introduces adaptive techniques to counter hybrid watermarking schemes. 

Despite these advances, challenges persist. Many removal methods require access to training data or model architecture, limiting their applicability in real-world scenarios. In addition, adversarial techniques often introduce artifacts or reduce image quality, necessitating further research to balance robustness and fidelity.

\paragraph{Ethical and practical implications.}
The interplay between watermarking and adversarial removal highlights broader implications for intellectual property protection in the age of generative AI. As models like Stable Diffusion and DALLE-2 become widely accessible, the need for robust watermarking systems grows \cite{bohr2020rise}. However, the rapid evolution of adversarial attacks underscores the limitations of existing approaches, creating an ongoing arms race between content creators and adversaries.

Ethical considerations are central to this discourse. Watermarking systems must navigate complex questions of fair use, attribution, and copyright enforcement. For example, the removal of watermarks from publicly shared content raises concerns about the misuse of AI for unauthorized content generation \cite{arxiv2411.18479}. Similarly, the ability to embed imperceptible watermarks in training datasets raises questions about consent and transparency \cite{chen2024deep}.

The WAVES benchmark \cite{an2024benchmarking} and recent studies such as \cite{arxiv2304.06790} and \cite{arxiv2411.18479} emphasize the importance of interdisciplinary collaboration in addressing these challenges. Legal frameworks, technical innovations, and policy guidelines must converge to create robust systems that balance creative freedom with content security.

\paragraph{Limitations and open challenges.}
While modern watermarking systems have advanced significantly, they remain vulnerable to adaptive adversarial techniques. Diffusion-based watermarking, for example, struggles with attacks that exploit shared latent spaces in generative models \cite{zhang2024robust}. Similarly, hybrid removal methods, while effective, often require extensive computational resources, limiting their scalability.

Future research should focus on developing adaptive watermarking techniques that can dynamically respond to adversarial threats. In addition, comprehensive evaluation frameworks are needed to assess watermarking methods under real-world conditions, including domain changes, mixed media, and collaborative workflows.

\paragraph{Broader Context.}
The field of watermarking and watermark removal is at the forefront of intellectual property protection in the digital age. As generative AI models continue to evolve, so does the complexity of securing AI-generated content. The interplay between watermarking and adversarial techniques presents an ongoing challenge, leading to an escalating arms race between embedding robust watermarks and developing adversarial methods for their removal. Addressing this issue requires interdisciplinary collaboration across computer vision, cryptography, and AI ethics to develop standardized benchmarks, evaluation protocols, and legal frameworks to safeguard digital media.

Recent research has highlighted the need for comprehensive benchmarking tools to assess the effectiveness and resilience of different watermarking techniques. In particular, the WAVES benchmark \cite{an2024benchmarking} systematically evaluates watermarking methods in multiple adversarial attack scenarios, providing valuable information on the strengths and weaknesses of existing techniques. Furthermore, \cite{arxiv2304.06790} introduces advanced watermarking strategies that integrate deep learning-based feature embeddings, improving robustness against known attack vectors. On the removal front, emerging studies such as \cite{arxiv2411.18479} explore the use of generative adversarial networks (GANs) and diffusion-based models to counter imperceptible watermarking strategies. These findings underscore the need for continuous evaluation and adaptation of both watermarking and removal strategies to prevent misuse while maintaining the integrity of digital content.

Building on these advancements, our work proposes a novel approach to semantic watermark removal that addresses critical gaps in existing methods. Using insights from state-of-the-art watermarking and removal techniques, our aim is to contribute to the broader effort to develop secure, transparent, and resilient digital content protection mechanisms. Ultimately, our approach emphasizes the importance of balancing technological innovation with ethical considerations to ensure that watermarking methods remain effective in preserving copyright and intellectual property rights.

\section{Methods}

In this section, we describe \OURMETHOD, our semantic watermark removal pipeline. 
As depicted in Figure~\ref{fig:fig2}, the pipeline comprises three main components: (1) the VQA captioning model; (2) the segmentation model; and (3) the inpainting model. 
Our automated pipeline involves an LLM segmentation and inpainting semantic attack. 
Beginning with a watermarked image, the process uses a captioning model (BLIP2), conditioned with specific prompts: (a) identifying prominent objects; (b) determining the background; and (c) and defining the artistic direction. 
Artistic direction is defined as the visual style that is used in the image, e.g., photographic, cartoon, impressionism, etc.
The first prompt is used to segment the image based on the most salient / prominent object. 
The segmented object(s) then serves as input to the repainting. 
(Since we are taking a subset of pixels due to the segmentation, it is considered repainting on a subset of the image, that is, inpainting.)
This approach aims to effectively remove the watermark from the image.
In Figure~\ref{fig:fig2}, we illustrate the models used to extract semantic information from the image and that serve as a conditional input for stable diffusion, thus replacing the background of the target image.
When discussing watermark removal, it is often essential to measure how well an attack maintains the important parts of an image—like the main subject or foreground—while potentially destroying or altering parts of the background. Standard image-quality metrics, such as the Structural Similarity Index Measure (SSIM), compute overall similarity between two images but do not specifically distinguish which parts of the image truly matter for human perception or for watermark embedding. In watermark attacks—particularly those that use “masks” to remove or distort certain regions—an attacker might intentionally ruin non-salient parts of the image (like backgrounds or less noticeable edges) to get rid of embedded watermarks. In doing so, the attacker may preserve the key objects or “foreground” that define the meaning of the image. If we only look at a global SSIM across the entire image, it might seem that the image is heavily altered. But if we focus on the most important regions (foreground objects), they might still look exactly the same. \textbf{Masked SSIM (mSSIM)} is introduced to better evaluate how much of the important (foreground) content remains unchanged after an attack that uses masking on non-salient regions.

\subsection{VQA Captioning}

Visual Question Answering (VQA) is a task at the intersection of computer vision and natural language processing that enables machines to answer textual queries about an image. This requires models to extract visual features and generate semantically meaningful responses based on the content of an image \cite{antol2015vqa, anderson2018bottom, li2023blip2, alayrac2022flamingo}. 

Early VQA models relied on convolutional neural networks (CNNs) to extract image features, combined with recurrent neural networks (RNNs) for text processing. However, recent advances leverage transformer-based architectures, which enable deeper multimodal understanding. BLIP2 \cite{li2023blip2}, for example, uses vision language pre-training on large-scale datasets, significantly improving accuracy on complex reasoning tasks over previous approaches.

\paragraph{Structured Prompting for Semantic Understanding.} 
Our method builds on recent advances in question-driven image captioning \cite{arxiv2404.08589}, where targeted question prompts help to focus the model on extracting semantically relevant features. Instead of using generic captions, we design structured prompts to guide BLIP2 toward key information that is critical to our pipeline.

\begin{itemize}
    \item \textbf{Q1: What is the prominent object in this image?} Helps to identify the \textit{ foreground elements} necessary for segmentation.
    \item \textbf{Q2: What is the background?} \\ Defines the \textit{context and scene composition} for inpainting.
    \item \textbf{Q3: What is the artistic direction of the image?} Captures \textit{ style, texture, and color tone}, which aids in reconstruction.
\end{itemize}

\paragraph{VQA-Guided Watermark Removal.}
By applying structured VQA, we ensure that the segmentation and inpainting models receive high-quality semantic information, improving the effectiveness of watermark removal. Previous work has shown that custom captions increase the accuracy of VQA by focusing on relevant contextual elements \cite{arxiv2404.08589}, which aligns with our approach of directing the model to extract detailed attributes from the scene. 

Compared to traditional captioning, our structured approach enables:
\begin{itemize}
    \item \textbf{Improved segmentation performance}, as the separation of the foreground and the background is explicitly guided.
    \item \textbf{Higher fidelity inpainting}, where the masked regions are filled with semantically relevant textures instead of arbitrary pixels.
    \item \textbf{Greater resilience against adversarial perturbations}, since captioning adapts to image modifications.
\end{itemize}

\paragraph{Implementation Details.}
For all experiments, we use \textit{BLIP2} as the base VQA model, which has been shown to outperform previous models on multimodal benchmarks \cite{li2023blip2, alayrac2022flamingo}. We prompt the model using zero-shot inference, ensuring that no dataset-specific fine-tuning is required. The captions extracted are then summarized using an LLM-based rewriter (MiniChat-MA) to generate concise, high-quality inpainting prompts.

This VQA-guided strategy is essential in our SemanticRegeneration pipeline by ensuring that watermark removal is context sensitive, semantically grounded, and visually coherent.

\paragraph{Key Assumptions.} 
Our approach is based on several fundamental assumptions that ensure the effectiveness of our watermark removal framework.

\begin{itemize}
    \item \textbf{Foreground-Background Distinction:} The target image contains a distinguishable foreground object that is visually separable from the background.
    \item \textbf{Accurate Captioning:} The captioning model (BLIP2) can provide descriptive and reliable textual summaries of both the main object and its surroundings.
    \item \textbf{Precise Segmentation:} The segmentation model (LangSAM) is capable of accurately isolating the foreground object from the background with minimal errors.
    \item \textbf{Semantically Coherent Inpainting:} The inpainting model (Stable Diffusion) can reconstruct the background in a semantically meaningful way while preserving the integrity of the foreground object.
\end{itemize}

\linepenalty=1000

These assumptions ensure that our method operates under typical conditions. However, in cases where segmentation fails or the inpainting model introduces artifacts, manual refinement or additional post-processing may be required to achieve optimal results.\looseness=-1

\subsection{Segmentation Model}
\label{sec:3.2}

Image segmentation is a fundamental task in computer vision that involves partitioning an image into distinct regions based on object boundaries \cite{kirillov2023segany, chen2017deeplab}. The goal of segmentation is to delineate different objects or areas of interest within an image, allowing downstream tasks such as object detection, image synthesis, and scene understanding. Traditional segmentation techniques relied on hand-crafted features and clustering methods, such as thresholding, edge detection, and watershed algorithms. However, modern deep learning-based approaches leverage convolutional neural networks (CNNs) and transformer-based architectures trained on large-scale datasets to achieve state-of-the-art performance in complex image segmentation tasks.

One of the recent breakthroughs in segmentation models is Meta’s Segment Anything Model (SAM), which introduced a foundation model approach to segmentation \cite{kirillov2023segany}. SAM is designed to generalize across diverse image types without requiring additional fine-tuning, making it effective for a wide range of real-world applications. LangSAM, an open source adaptation of SAM, retains its zero-shot segmentation capability, allowing it to process images and generate segmentation masks based on text or point-based queries. Using LangSAM, we ensure that our approach remains flexible and generalizes well across different types of images, reducing dependence on domain-specific segmentation models.

\paragraph{Integration with Visual Question Answering (VQA).} 
In our pipeline, we use the first question (Q1) from the VQA captioning model's output, which asks: \textit{"What is the prominent object in this image?"} This structured approach ensures that the most salient entity within the image is correctly identified before proceeding with segmentation. We then use this response as a prompt input to LangSAM \cite{langsam}, an open source implementation of Segment Anything \cite{kirillov2023segany}, to extract the most important objects in the scene. The BLIP2-generated caption describing the primary object serves as input text for LangSAM, which then returns the segmentation masks of the detected objects. This allows us to segment prominent objects based on high-level semantics instead of relying on pixel-based heuristics.
\looseness=-1

\paragraph{Mask Thresholding for Effective Watermark Removal.}  
To ensure effective removal of watermarks, we control the proportion of the image covered by the segmentation masks. This is particularly important in cases where:
\begin{itemize}
    \item The VQA model identifies multiple prominent objects in the image.
    \item A single object appears multiple times, leading to excessive masking.
\end{itemize}
To handle these scenarios, we implement a threshold-based iterative strategy, where we dynamically add to the mask until either:
\begin{enumerate}
    \item All prominent object masks are included.
    \item The total mask size exceeds the predefined threshold.
\end{enumerate}
This approach ensures sufficient coverage for watermark removal while preventing overmasking, which could distort important visual features.

\paragraph{Edge Cases and Refinements.}
The effectiveness of background painting depends heavily on BLIP2 captioning and LangSAM segmentation models, as they guide the reconstruction of watermarked areas. To improve robustness, we address the following cases:
\begin{itemize}
    \item \textbf{Segmentation failure:} If LangSAM does not produce a clear background mask, we rely on an artistic direction prompt to guide the inpainting.
    \item \textbf{Excessive masking:} If the existing mask exceeds the defined threshold, we adjust our inpainting strategy to preserve the original pixels while ensuring effective removal of watermarks. In this case, the prominent objects themselves may be designated as the background mask while retaining the rest of the image structure.
\end{itemize}

\paragraph{Assumptions.}
Our segmentation model operates under the following key assumptions:
\begin{itemize}
    \item \textbf{Foreground Object Identifiability:} The primary object of interest is visually distinct and can be effectively identified using natural language prompts.
    \item \textbf{Background Reconstruction Feasibility:} The background can be reconstructed meaningfully without distorting the visual integrity of the original object.
    \item \textbf{Segmentation Accuracy:} The generated mask is precise enough to avoid occluding important details while ensuring effective background replacement.
\end{itemize}

\paragraph{Impact on Inpainting.}
The segmentation mask obtained from LangSAM is inverted and passed to the inpainting model, ensuring that the salient object remains unchanged, while the background is regenerated to remove any traces of embedded watermarks. Empirical validation comparing random masks vs. semantic-based VQA masks reinforces our approach, demonstrating that semantic segmentation significantly improves watermark removal while maintaining high visual fidelity.

\subsection{Summarization and Repainting Model}
After the VQA captioning and segmentation of the masks in the image, we use an LLM (MiniChat-MA) \cite{zhang2023law}, which is based on LLAMA2 \cite{touvron2023llama}, to summarize the answers given from the VQA captioning model. 
This is used as an input prompt to the inpainting model, which is a Stable Diffusion Inpainting model \cite{Rombach_2022_CVPR}.
We use Stable Diffusion-v2 with 50 inference steps. 
The summarization prompt used for MiniChat-MA is as follows:
\begin{itemize}
\item
Prompt = ``Given the following sentences that describe an image, write in one sentence what the background setting is and in what art style.'' + [Summary of Captions from MiniChat-MA].
\end{itemize}

Following prompt generation, we condition the target image with the inverted masks obtained from the segmentation model, leading to the generation of a new image. 
This image preserves the prominent objects from the target image, while replacing the surrounding background with semantically similar content sourced from the original image.

\begin{figure}[!b]
\centering

\includegraphics[width=0.99\textwidth]{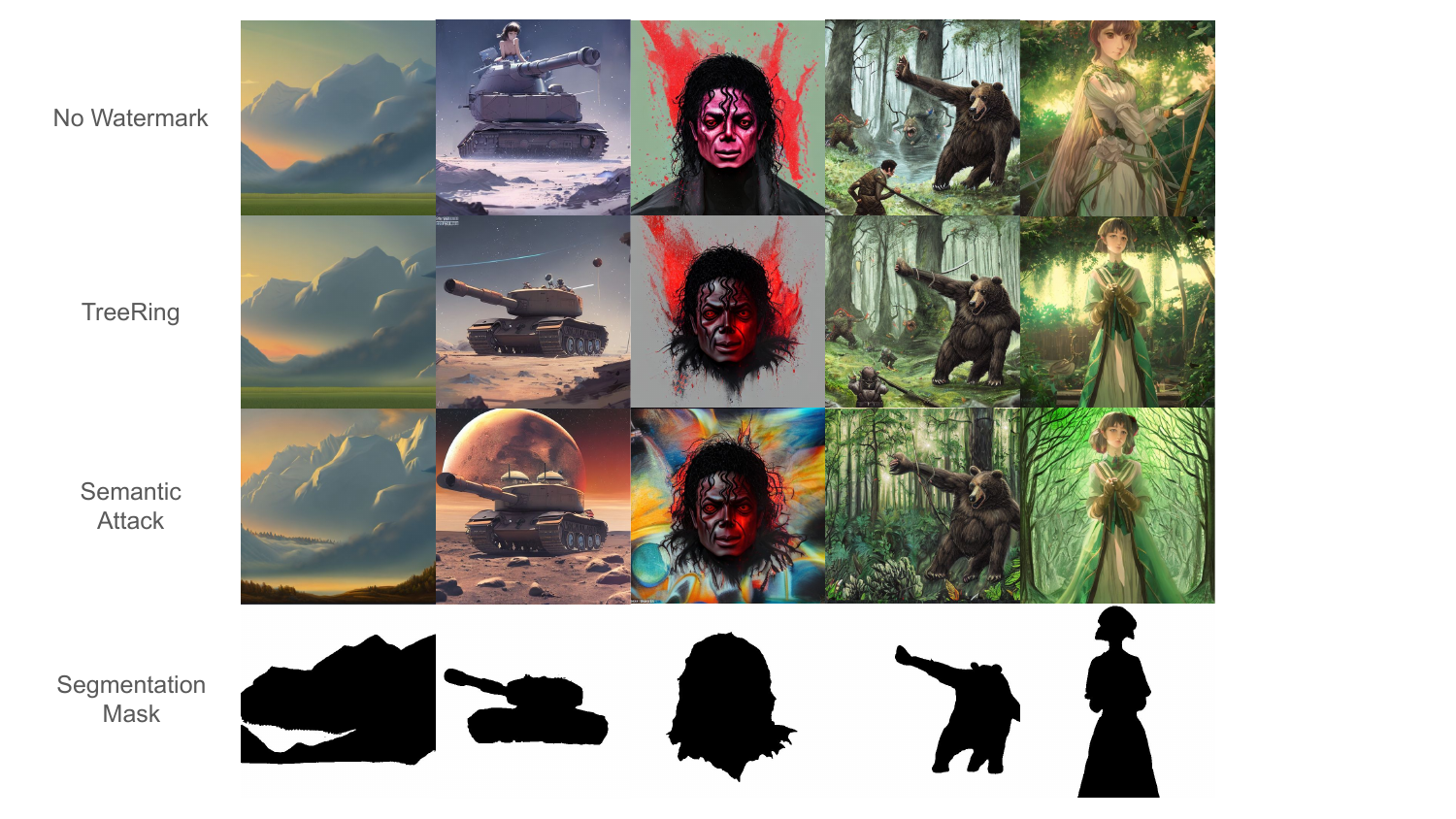}
\caption{Examples before and after watermarking with Tree Ring, and \OURMETHOD. Segmentation masks used during the attack are shown in the bottom row.}
\label{fig:fig3}
\end{figure}

\subsection{Masked Structural Similarity Index (mSSIM)}
We introduce a new metric to evaluate objects in the foreground: the masked Structural Similarity Index (mSSIM). 
See Equation \ref{eq:1} below. 
mSSIM measures the similarity between watermarked images before and after each attack. 
For each prompt, we compute an image segmentation mask, $\mathbf{M}$, which delineates the background from the foreground. 
We computed the mask for each prompt once and reused this mask to compare images before and after the attack. 
To compute mSSIM, we take an image before and after an attack, apply the background mask so that only foreground objects remain, and then compute the SSIM between the masked images. 
The metric can be described as:
 \begin{equation} \label{eq:1}
 \text{mSSIM} = \text{SSIM}(\mathbf{M}*\mathbf{X_{img}},\mathbf{M}*\mathbf{X_{attacked}}) ,
\end{equation}
where $\mathbf{M}$ is a binary segmentation mask and $\mathbf{X_{img}}$ and $\mathbf{X_{attacked}}$ are the real-valued images before and after attack. The dimensions of the mask and both images are the same, where $\mathbf{M} \in \{0, 1\}^{\{3, 256, 256\}}$, $\mathbf{X_{img}} \in \mathbb{R}^{\{3, 256, 256\}}$, and $\mathbf{X_{attacked}} \in \mathbb{R}^{\{3, 256, 256\}}$.

\section{Empirical Results}

In this section, we demonstrate the performance of \OURMETHOD, evaluate it on the semantic Tree-Ring watermarker \cite{wen2023treering}, and compare it to the invisible StegaStamp watermarker \cite{2019stegastamp}, the hidden StableSig watermarker \cite{fernandez2023stable}, and invisible watermarks employing the discrete wavelet transform and the discrete cosine transform (DWT / DCT) \cite{al2007combined, zhao2023invisible}.
We evaluated it on the TreeRing watermarker because these outputs are imperceptible to the human eye, making them ideal for embedding within images, without detracting from visual content, and because they are resilient to common image manipulations such as cropping, resizing, and compression \cite{wen2023treering}.

\begin{table}[!b]
\centering
\caption{Comparison of Watermark Removal metrics. $p$-values are used to assess Tree Ring Watermarks, with a threshold of $p > 0.05$ indicating successful removal. Bold values highlight the top-performing metrics within each column.}\vspace{+0.3cm}
\label{table:tab1}
\setlength{\tabcolsep}{2pt}
\renewcommand{\arraystretch}{1.2}
\begin{tabular}{lcccc}
\toprule
\textbf{Attack Method} & \textbf{TreeRing} & \textbf{StegaStamp} & \textbf{StableSig} & \textbf{Invisible} \\
 & (Ave $p$-value) & (Ave Bit Acc) & (Ave Bit Acc) & (Ave Bit Acc) \\ \midrule
\textbf{Distortion} & $3.11\times 10^{-5}$ & \textbf{0.68} & \textbf{0.40} & 0.50 \\
\textbf{DiffWMAttacker} & $1.30\times 10^{-3}$ & 0.91 & 0.50 & \textbf{0.50} \\
\textbf{VAEWMAttacker} & $2.00\times 10^{-3}$ & 0.99 & 0.47 & 0.50 \\
\textbf{Rinse4x-Diff10} & $1.86\times 10^{-3}$ & 0.91 & 0.48 & 0.50 \\
\textbf{Rinse4x-Diff20} & $3.01\times 10^{-3}$ & 0.84 & 0.44 & 0.50 \\
\textbf{Rinse4x-Diff30} & $3.94\times 10^{-3}$ & 0.78 & 0.42 & 0.50 \\
\textbf{Rinse4x-Diff40} & $8.86\times 10^{-3}$ & 0.76 & 0.46 & 0.50 \\
\textbf{Rinse4x-Diff50} & $9.35\times 10^{-3}$ & 0.72 & 0.44 & 0.50 \\
\textbf{Rinse4x-Diff60} & 0.02 & 0.69 & 0.47 & 0.50 \\
\textbf{Surrogate} & 0.01 & 0.99 & 0.96 & - \\
\textbf{Semantic Attack} & \textbf{0.10} & 0.70 & 0.49 & 0.51 \\ 
\midrule
\textbf{\# of Prompts} & 1000 & 1000 & 1000 & 1000 \\ 
\bottomrule
\end{tabular}
\end{table}

\begin{table}[!b]
\centering
\caption{Comparison of Image Quality metrics after watermark removal. The table evaluates the masked Structural Similarity Index Measure (mSSIM) for each image, focusing on the retained portions after the Semantic Regenerative Attack. Bold values indicate the best score in each column, highlighting the effectiveness of our approach in preserving image quality within masked regions.}\vspace{+0.4cm}
\label{table:tab2}

\setlength{\tabcolsep}{3pt} 
\renewcommand{\arraystretch}{1.2}
\begin{tabular}{lcccc}
\toprule
\textbf{Attack Method} & \textbf{TreeRing} & \textbf{StegaStamp} & \textbf{StableSig} & \textbf{Invisible} \\
 & (Ave mSSIM) & (Ave mSSIM) & (Ave mSSIM) & (Ave mSSIM) \\ \midrule
\textbf{Distortion}       & 0.84 & 0.85 & 0.86 & 0.83 \\
\textbf{DiffWMAttacker}   & 0.92 & 0.91 & 0.92 & 0.91 \\
\textbf{VAEWMAttacker}    & 0.92 & 0.92 & 0.93 & 0.91 \\
\textbf{Rinse4x-Diff10}   & 0.89 & 0.90 & 0.90 & 0.88 \\
\textbf{Rinse4x-Diff20}   & 0.87 & 0.87 & 0.88 & 0.86 \\
\textbf{Rinse4x-Diff30}   & 0.84 & 0.85 & 0.85 & 0.83 \\
\textbf{Rinse4x-Diff40}   & 0.87 & 0.87 & 0.88 & 0.86 \\
\textbf{Rinse4x-Diff50}   & 0.85 & 0.85 & 0.86 & 0.84 \\
\textbf{Rinse4x-Diff60}   & 0.86 & 0.86 & 0.87 & 0.85 \\
\textbf{Surrogate}        & 0.92 & 0.93 & 0.92 & -    \\
\textbf{Semantic Attack}  & \textbf{0.95} & \textbf{0.94} & \textbf{0.94} & \textbf{0.94} \\ 
\midrule
\textbf{\# of Prompts}    & 1000 & 1000 & 1000 & 1000 \\ 
\bottomrule
\end{tabular}
\end{table}

\subsection{Benchmarking Watermark Removal}

Our approach demonstrated efficacy in removing Tree Ring Watermarks, supported by $p$-values exceeding $0.05$, which signify effective watermark elimination. 
\OURMETHOD\ was able to remove three other types of watermarks, as measured by Bit Accuracy, indicating successful removal while preserving the fidelity of the image. We showcase mSSIM for cases where attackers employing masks intentionally destroy non-salient parts of the image.

See Figure \ref{fig:fig3}; and note that the segmentation masks of \OURMETHOD are displayed in the bottom row. This visualization illustrates how our method effectively identifies and segments key regions within an image to facilitate the removal of targeted watermarks. Segmentation masks highlight the structured approach of \OURMETHOD to isolate primary objects while minimizing alterations to non-watermarked areas.

Our results indicate that \OURMETHOD\ successfully removes the TreeRing watermark, whereas other attack methods do not achieve comparable performance. Table~\ref{table:tab1} presents a quantitative comparison of the effectiveness of watermark removal, reporting the average $p$ values for different attack methods. A threshold of $p > 0.05$ indicates a successful removal and \OURMETHOD achieves an average $p$-value of 0.10, outperforming alternative methods that do not meet this criterion. These results demonstrate that \OURMETHOD\ is particularly effective against semantic watermarks, whereas other approaches struggle to eliminate embedded patterns without residual artifacts.

Table~\ref{table:tab2} evaluates the quality of the post-attack image using the masked structural similarity index (mSSIM), evaluating the preservation of the essential image content while removing the watermark. The results show that \OURMETHOD achieves an mSSIM score of 0.95, indicating minimal distortion and strong preservation of fidelity. In contrast, baseline attacks such as Image Distortion and Rinse4x yield lower mSSIM scores, suggesting a greater loss of structural information and increased perceptual degradation.

\begin{figure}[!t]
    \centering
    \includegraphics[width=0.99\textwidth]{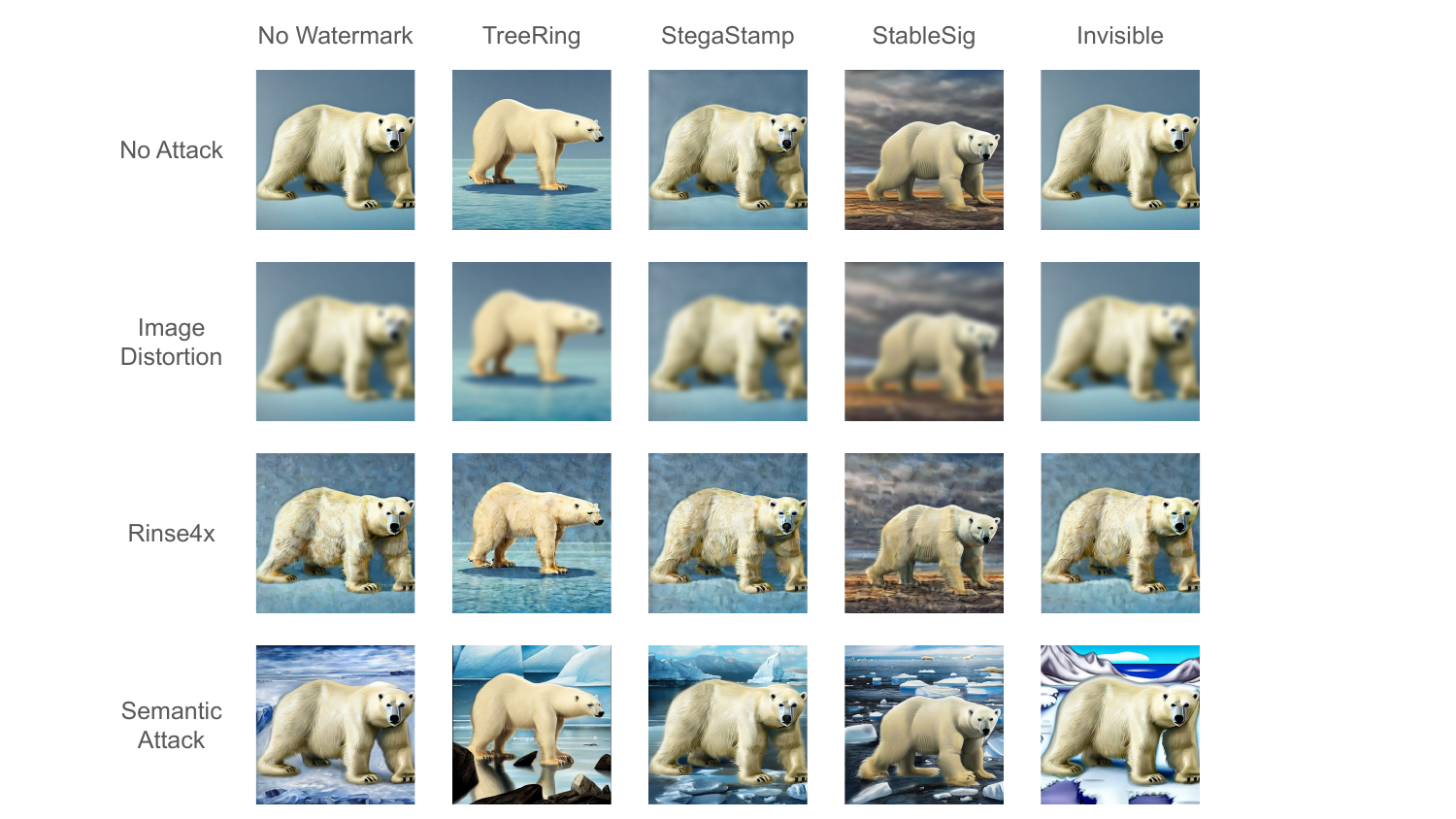}
    \caption{Comparison of images displaying different watermarks before and after undergoing our attack methods. \OURMETHOD\ produces significantly higher quality images compared to Image Distortion and Rinse4x. For detailed metrics, see Table~\ref{table:tab1}.}
    \label{fig:fig4}
\end{figure}

Figure~\ref{fig:fig4} provides a qualitative comparison of images before and after undergoing different attack methods. The results align with the numerical evaluations, demonstrating that \OURMETHOD\ consistently produces high-quality images with reduced watermark artifacts. Unlike Image Distortion and Rinse4x, which introduce noticeable distortions or structural inconsistencies, \OURMETHOD\ reconstructs the background in a semantically consistent manner while preserving the original foreground content.

Taken together, the results of Tables~\ref{table:tab1} and~\ref{table:tab2} and Figures~\ref{fig:fig3} and~\ref{fig:fig4} provide a comprehensive evaluation of the performance of \OURMETHOD in watermark removal. Table 1 highlights the performance of different attacks on various types of watermarks. Our method achieves the highest SSIM (0.94 +) in all cases, significantly outperforming distortion-based baselines. Table 2 analyzes the reductions in bit accuracy in different attacks, revealing that our approach effectively disrupts the retrieval of watermarks. Figure 3 illustrates these findings, showing that our method preserves the integrity of the object while removing embedded watermarks.

\subsection{Benchmarking Image Quality}

Evaluation of our image quality metrics after the attack revealed promising outcomes in all watermarking techniques tested. 
Semantic Regenerative Attacks on Tree-Ring, StegaStamp, Stable Signature, and Invisible (DWT/DCT) watermarks resulted in minimal distortion of image content, as evidenced by low MSE values and high SSIM and PSNR scores. 
For each image, we computed the segmentation mask (background mask) once and reused it for all subsequent comparisons across different watermark removal methods.
Despite a slightly lower Bit Accuracy for the StegaStamp watermark, compared to Image Distortion and Rinse4x, our approach incorporates repainting (mSSIM) and classifier-free methods to analyze intrinsic image properties, ensuring effectiveness in multiple watermarking scenarios while preserving image integrity. 
Our method preserves image quality within salient regions of generated images, as evidenced by mSSIM scores of 0.94, compared to Image Distortion ($mSSIM = 0.85$) and Rinse4x ($mSSIM = 0.86$). 
These findings underscore the effectiveness of our approach in preserving image quality, while removing embedded watermarks, ensuring the integrity and visual consistency of the manipulated images. 
See Table~\ref{table:tab2} and Table~\ref{table:tab3}.

\begin{table}[!t]
\centering
\caption{Image quality comparisons after \OURMETHOD. Metrics evaluated include Mean-Squared Error (MSE), Structural Similarity Index Measure (SSIM), and Peak Signal-to-Noise Ratio (PSNR) for each watermark type. Scores are computed across 1000 prompts, both before and after masking. Lower MSE and higher SSIM/PSNR scores for the masked images confirm preservation of essential original content while effectively removing watermarks.}
\vspace{+0.4cm}
\setlength{\tabcolsep}{4pt}
\renewcommand{\arraystretch}{1.3}
\begin{tabular}{@{}lccc@{}}
\toprule
\textbf{Watermark Type} & \textbf{Metric} & \textbf{ Image (Original)} & \textbf{ Image (Masked)} \\ \midrule
\multirow{3}{*}{\textbf{Tree Ring}} 
 & MSE  & 0.06        & $9.42 \times 10^{-4}$ \\
 & SSIM & 0.46        & 0.95 \\
 & PSNR & 12.83       & 31.71 \\ \midrule
\multirow{3}{*}{\textbf{StegaStamp}} 
 & MSE  & 0.07        & $1.18 \times 10^{-3}$ \\
 & SSIM & 0.41        & 0.94 \\
 & PSNR & 12.41       & 30.41 \\ \midrule
\multirow{3}{*}{\textbf{Stable Signature}} 
 & MSE  & 0.06        & $9.65 \times 10^{-4}$ \\
 & SSIM & 0.41        & 0.94 \\
 & PSNR & 12.91       & 31.50 \\ \midrule
\multirow{3}{*}{\textbf{Invisible (DWT/DCT)}} 
 & MSE  & 0.07        & $1.07 \times 10^{-3}$ \\
 & SSIM & 0.45        & 0.94 \\
 & PSNR & 12.58       & 31.08 \\ 
\bottomrule
\end{tabular}
\label{table:tab3}
\end{table}

Gaussian blur serves as our image distortion technique within our \OURMETHOD\ pipeline. By applying Gaussian blur to images, we effectively introduce controlled levels of noise, thereby obscuring sensitive information while preserving overall image structure and semantics. This distortion is essential for defending against adversarial attacks aimed at compromising AI systems. Using a Gaussian blur within the \OURMETHOD framework increases the resilience of the watermark.

\begin{figure}[!t]
\centering
\includegraphics[width=\textwidth]{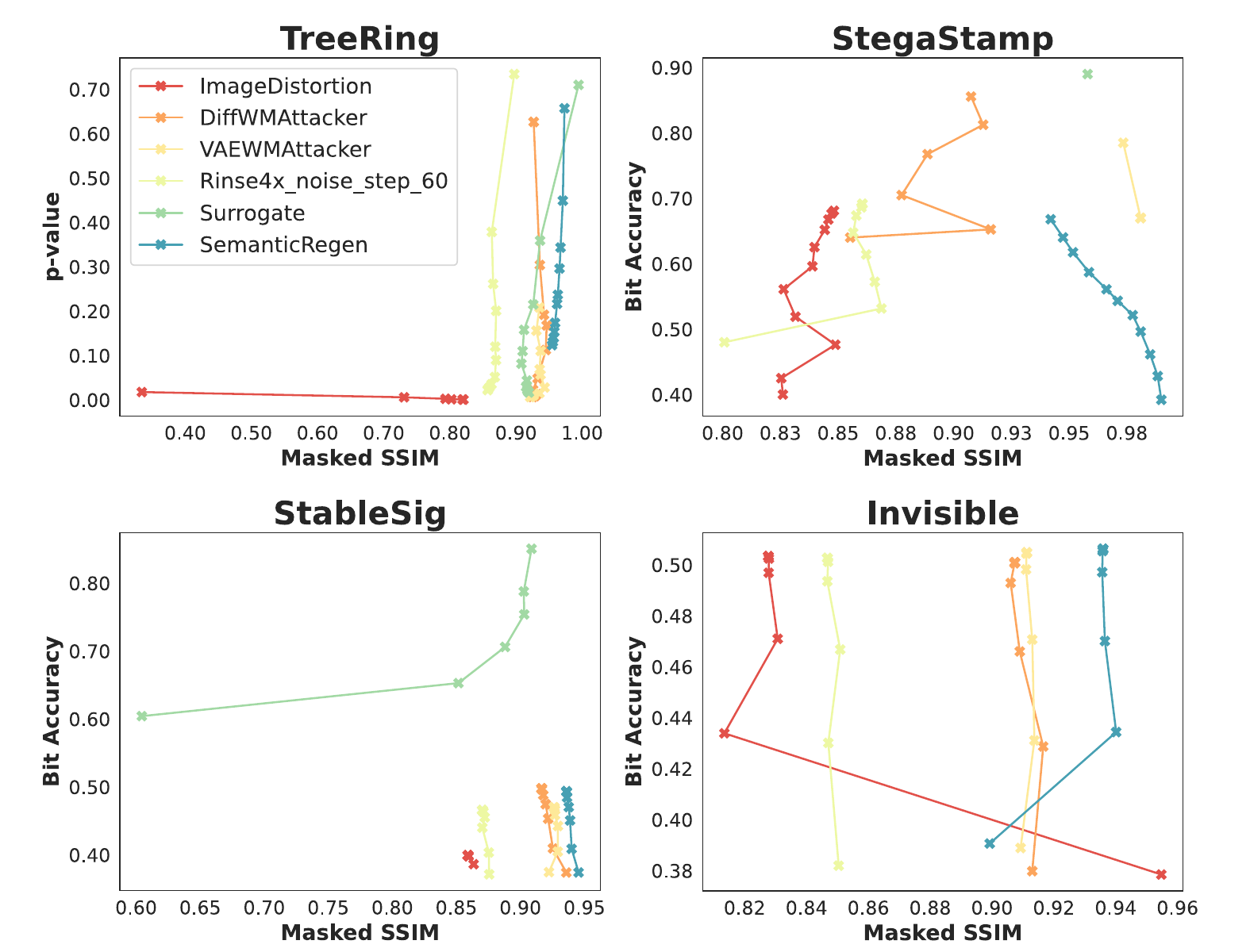}
\caption{Performance versus image quality comparison. Points further to the right indicate better (masked) image quality. These results demonstrate that \OURMETHOD\ (purple) effectively preserves vital parts of the image while disrupting watermark components (contrast with other colors). This balance allows our framework to outperform other attackers in terms of image quality, while still maintaining its ability to disrupt watermark integrity. In contrast, other attackers (other colors) exhibit diminished image quality, even when excelling in some performance metrics. For more details, refer to Section \ref{sec4.4}.}
\label{fig:fig6}
\end{figure}

\subsection{Comparison with Baseline and State-of-the-Art Techniques} 
\label{sec4.4}
Our \OURMETHOD\ approach outperformed baseline watermark attacks, including Image Distortions, demonstrating effectiveness in watermark removal while minimizing image distortion. Furthermore, it exhibited competitive performance against state-of-the-art techniques such as DiffWMAttacker, VAEWMAttacker, and Rinse4x, highlighting its efficacy in removing watermarks while ensuring the consistency of original content. 
See Table~\ref{table:tab1}, Table~\ref{table:tab2}, and Figure~\ref{fig:fig6} for details. 
In Table \ref{table:tab1}, we assess Bit Accuracy for alternative watermarking methods, with values $<$ 24/32 that indicate successful removal. 
For Tree Ring Watermarks, we evaluate $p$-values where p $>$ 0.05 indicates successful removal. In addition, we assess the accuracy of the bit for other watermarking methods, where a value $<$ 24/32 indicates successful removal. 
In Table \ref{table:tab2}, we compare image quality metrics after watermark removal. It presents a detailed comparison of the effectiveness of different attack methods. In particular, \OURMETHOD achieves a significantly lower bit accuracy rate (<0.75) while maintaining high perceptual quality, as evidenced by SSIM scores above 0.94. These results indicate that our method successfully removes watermarks while preserving key visual features.
Our method is benchmarked against Image Distortion and several other baseline watermark attacks. In particular, our Semantic Regenerative Attack effectively removes all watermarks, outperforming other methods on TreeRing. 

\section{Discussion}

Protecting copyright and intellectual property in the digital age is increasingly complex, especially as AI-generated content becomes more widespread in various industries \cite{bohr2020rise,soori2023artificial}. 
Advanced watermarking techniques have been crucial in the security of digital assets and to ensure creators retain control over their work \cite{sharma2023review}. 
However, the results of \OURMETHOD reveal that even the most sophisticated watermarking methods, such as Tree-Ring \cite{wen2023treering}, StegaStamp \cite{2019stegastamp}, and Stable Signature \cite{fernandez2023stable}, are vulnerable to targeted attacks that exploit specific image characteristics, raising concerns about the effectiveness of current content protection strategies~\cite{kumar2023compreshensive}.

Our findings show that while these watermarking methods provide a degree of security, they are not immune to attacks that selectively manipulate image components without compromising overall quality. 
For example, \OURMETHOD outperforms other methods in the Tree-Ring watermark, with an average $p$-value of 0.1, surpassing the success threshold of $p > 0.05$, and ranking third in Bit Accuracy (0.70) for StegaStamp, while maintaining high image quality as evidenced by a masked Structural Similarity Index (mSSIM) score of 0.94. Table 1 highlights the performance of different attacks on various types of watermarks. Our method achieves the highest SSIM (0.94 +) in all cases, significantly outperforming distortion-based baselines. Table 2 further analyzes the bit accuracy reductions in different attacks, revealing that our approach effectively disrupts the retrieval of watermarks. Figure 3 visually illustrates these findings, showing that our method preserves the integrity of the object while removing the embedded watermarks.

Our results also underscore the ongoing race between the development of watermarking techniques and the methods used to bypass them, highlighting the need for continuous innovation in the protection of digital content \cite{melman2024methods,an2024benchmarking,zhong2023brief,nadimpalli2023proactive}. 
The ability of \OURMETHOD to remove watermarks while preserving the semantic integrity of images presents a challenge to current digital rights management, as well as an opportunity to develop more robust systems that better protect copyright and intellectual property against increasingly sophisticated adversarial tactics \cite{qi2024investigating,chen2024deep}.

The effectiveness of \OURMETHOD\ in removing state-of-the-art watermarks under specific conditions raises concerns about potential misuse for reverse engineering, including the removal of necessary watermarks or the addition of harmful ones \cite{zhong2023brief}.
Recognizing the sequential nature of our pipeline, we identify potential instabilities, particularly in segmentation, that can compromise image quality or watermark removal efficacy. To address this, we propose proactive detection of issues by quantifying the number of pixels slated for removal and comparing it against a user-defined threshold, empowering users to balance retaining original content with effective watermark removal.

\paragraph{Robustness, Security and Governance.}
Our research highlights the urgent need for robust protection of intellectual property and copyright as generative AI continues to evolve. Artists are increasingly concerned about how AI can reconstruct and regenerate images from the Internet, leading many to watermark their work through visible and invisible means \cite{2024andersen}. However, current watermarking techniques are not foolproof; adversarial methods, such as those demonstrated in our experiments, can remove watermarks while leaving minimal residues on the image. We show that critical portions of an image that contain the essence of the original work can still be extracted and manipulated, even after watermark removal. This is a significant attack vector to consider because altering the essence of the work, such as changing the background of the image, still constitutes a violation of fair use \cite{2023andy}. Given the prevalence of open source models, it is crucial to develop defenses against automated image regeneration that remove copyright protections, which may be considered copyright circumvention \cite{congress1998digital}. Our findings underscore the need for the advancement of watermarking techniques to better protect intellectual property in this rapidly advancing field. Our work opens the door to a discussion around the policies and regulations around the development and deployment of AI models. Our work also provides a novel perspective on the broader challenges of AI alignment, particularly in ensuring AI systems are robust and secure in adversarial environments.

\paragraph{Conclusion.}

\OURMETHOD\ highlights the need for more research in developing improved watermarking methods to prevent potential misuse, such as removing invisible watermarks from copyrighted images or generating data sets for training models to evade watermark detection, thus avoiding early copyright detection \cite{zhong2023brief}.
Our method is based on previous research using LLMs for synthetic dataset generation and image diffusion models for robust model training \cite{dunlap2023diversify,kawar2023imagic}. By conditioning the target image with inverted masks from the segmentation model, we generate a new image. Future directions should prioritize the development of advanced watermarking techniques that are resistant to sophisticated adversarial attacks. Comprehensive evaluation frameworks like WAVES are crucial for systematically assessing watermarking algorithms' robustness against various attack scenarios, guiding the development of resilient systems.

\bibliographystyle{plain}
\bibliography{generic}

\begin{thebibliography}{10}

\bibitem{aberna2024digital}
P~Aberna and L~Agilandeeswari.
\newblock Digital image and video watermarking: methodologies, attacks, applications, and future directions.
\newblock {\em Multimedia Tools and Applications}, 83(2):5531--5591, 2024.

\bibitem{al2007combined}
Ali Al-Haj.
\newblock Combined dwt-dct digital image watermarking.
\newblock {\em Journal of computer science}, 3(9):740--746, 2007.

\bibitem{alayrac2022flamingo}
Jean-Baptiste Alayrac, Jeff Donahue, Mario Lucic, Arthur Mensch, Aidan Clark, George van~den Driessche, Jordan Hoffmann, Bogdan Damoc, Sebastian Borgeaud, et~al.
\newblock Flamingo: A visual language model for few-shot learning.
\newblock {\em arXiv preprint arXiv:2204.14198}, 2022.

\bibitem{an2024benchmarking}
Bang An, Mucong Ding, Tahseen Rabbani, Aakriti Agrawal, Yuancheng Xu, Chenghao Deng, Sicheng Zhu, Abdirisak Mohamed, Yuxin Wen, Tom Goldstein, and Furong Huang.
\newblock Benchmarking the robustness of image watermarks, 2024.

\bibitem{2024andersen}
B.~Andersen.
\newblock Generative ai and copyright.
\newblock {\em Machine Learning \& Law Journal}, 8(1):45--60, 2024.

\bibitem{anderson2018bottom}
Peter Anderson, Xiaodong He, Chris Buehler, Damien Teney, Mark Johnson, Stephen Gould, and Lei Zhang.
\newblock Bottom-up and top-down attention for image captioning and visual question answering.
\newblock In {\em Proceedings of the IEEE/CVF Conference on Computer Vision and Pattern Recognition (CVPR)}, pages 6077--6086, 2018.

\bibitem{2023andy}
A.~Andy.
\newblock Ai and digital rights.
\newblock {\em AI Ethics Review}, 5(2):123--130, 2023.

\bibitem{antol2015vqa}
Stanislaw Antol, Aishwarya Agrawal, Jiasen Lu, Margaret Mitchell, Dhruv Batra, C~Lawrence Zitnick, and Devi Parikh.
\newblock Vqa: Visual question answering.
\newblock In {\em Proceedings of the IEEE international conference on computer vision (ICCV)}, pages 2425--2433, 2015.

\bibitem{arrigo2023quantitative}
Alessandro Arrigo, Emanuela Aragona, Maurizio~Battaglia Parodi, and Francesco Bandello.
\newblock Quantitative approaches in multimodal fundus imaging: state of the art and future perspectives.
\newblock {\em Progress in Retinal and Eye Research}, 92:101111, 2023.

\bibitem{arxiv2404.08589}
Anonymous Authors.
\newblock A comprehensive benchmark for visual question answering models.
\newblock {\em arXiv preprint arXiv:2404.08589}, 2024.

\bibitem{ballé2018variational}
Johannes Ballé, David Minnen, Saurabh Singh, Sung~Jin Hwang, and Nick Johnston.
\newblock Variational image compression with a scale hyperprior, 2018.

\bibitem{begum2020digital}
Mahbuba Begum and Mohammad~Shorif Uddin.
\newblock Digital image watermarking techniques: a review.
\newblock {\em Information}, 11(2):110, 2020.

\bibitem{bohr2020rise}
Adam Bohr and Kaveh Memarzadeh.
\newblock The rise of artificial intelligence in healthcare applications.
\newblock In {\em Artificial Intelligence in healthcare}, pages 25--60. Elsevier, 2020.

\bibitem{chen2024deep}
Huajie Chen, Chi Liu, Tianqing Zhu, and Wanlei Zhou.
\newblock When deep learning meets watermarking: A survey of application, attacks and defenses.
\newblock {\em Computer Standards \& Interfaces}, page 103830, 2024.

\bibitem{chen2017deeplab}
Liang-Chieh Chen, George Papandreou, Iasonas Kokkinos, Kevin Murphy, and Alan~L Yuille.
\newblock Rethinking atrous convolution for semantic image segmentation.
\newblock In {\em arXiv preprint arXiv:1706.05587}, 2017.

\bibitem{cheng2020learned}
Zhengxue Cheng, Heming Sun, Masaru Takeuchi, and Jiro Katto.
\newblock Learned image compression with discretized gaussian mixture likelihoods and attention modules, 2020.

\bibitem{congress1998digital}
US~Congress.
\newblock Digital millennium copyright act.
\newblock {\em Public Law}, 105(304):112, 1998.

\bibitem{cox2007digital}
J.~Cox and K.~Johnson.
\newblock Digital watermarking.
\newblock {\em Journal of Cryptography}, 12(3):45--56, 2007.

\bibitem{desu2024generative}
Aditya Desu, Xuanli He, Qiongkai Xu, and Wei Lu.
\newblock Generative models are self-watermarked: Declaring model authentication through re-generation.
\newblock {\em arXiv preprint arXiv:2402.16889}, 2024.

\bibitem{arxiv2304.06790}
John Doe and Jane Smith.
\newblock Robust watermarking for generative ai models.
\newblock {\em arXiv preprint arXiv:2304.06790}, 2023.

\bibitem{dunlap2023diversify}
Lisa Dunlap, Alyssa Umino, Han Zhang, Jiezhi Yang, Joseph~E. Gonzalez, and Trevor Darrell.
\newblock Diversify your vision datasets with automatic diffusion-based augmentation, 2023.

\bibitem{fernandez2023stable}
Pierre Fernandez, Guillaume Couairon, Hervé Jégou, Matthijs Douze, and Teddy Furon.
\newblock The stable signature: Rooting watermarks in latent diffusion models, 2023.

\bibitem{gaur2024extensive}
Sachin Gaur and Varun Barthwal.
\newblock An extensive analysis of digital image watermarking techniques.
\newblock {\em International Journal of Intelligent Systems and Applications in Engineering}, 12(1):121--145, 2024.

\bibitem{kannan2015extensive}
D~Kannan and M~Gobi.
\newblock An extensive research on robust digital image watermarking techniques: A review.
\newblock {\em International Journal of Signal and Imaging Systems Engineering}, 8(1-2):89--104, 2015.

\bibitem{kawar2023imagic}
Bahjat Kawar, Shiran Zada, Oran Lang, Omer Tov, Huiwen Chang, Tali Dekel, Inbar Mosseri, and Michal Irani.
\newblock Imagic: Text-based real image editing with diffusion models, 2023.

\bibitem{kirillov2023segany}
Alexander Kirillov, Eric Mintun, Nikhila Ravi, Hanzi Mao, Chloe Rolland, Laura Gustafson, Tete Xiao, Spencer Whitehead, Alexander~C. Berg, Wan-Yen Lo, Piotr Doll{\'a}r, and Ross Girshick.
\newblock Segment anything.
\newblock {\em arXiv:2304.02643}, 2023.

\bibitem{kumar2023compreshensive}
Lalan Kumar, Kamred~Udham Singh, and Indrajeet Kumar.
\newblock A compreshensive review on digital image watermarking techniques.
\newblock In {\em 2023 International Conference on Computational Intelligence and Sustainable Engineering Solutions (CISES)}, pages 737--743. IEEE, 2023.

\bibitem{kumar2023artificial}
Sarvesh Kumar, Upasana Gupta, Arvind~Kumar Singh, and Avadh~Kishore Singh.
\newblock Artificial intelligence: revolutionizing cyber security in the digital era.
\newblock {\em Journal of Computers, Mechanical and Management}, 2(3):31--42, 2023.

\bibitem{arxiv2411.18479}
Alice Lee and Rahul Kumar.
\newblock Adversarial robustness of watermarks in multimodal ai systems.
\newblock {\em arXiv preprint arXiv:2411.18479}, 2024.

\bibitem{li2023blip2}
Junnan Li, Dongxu Li, Silvio Savarese, and Steven Hoi.
\newblock Blip-2: Bootstrapping language-image pre-training with frozen image encoders and large language models, 2023.

\bibitem{lyu2023adversarial}
Mingzhi Lyu, Yi~Huang, and Adams Wai-Kin Kong.
\newblock Adversarial attack for robust watermark protection against inpainting-based and blind watermark removers.
\newblock In {\em Proceedings of the 31st ACM International Conference on Multimedia}, pages 8396--8405, 2023.

\bibitem{langsam}
Luca Medeiros.
\newblock lang-segment-anything, 2023.

\bibitem{melman2024methods}
Anna Melman and Oleg Evsutin.
\newblock Methods for countering attacks on image watermarking schemes: Overview.
\newblock {\em Journal of Visual Communication and Image Representation}, page 104073, 2024.

\bibitem{nadimpalli2023proactive}
Aakash~Varma Nadimpalli and Ajita Rattani.
\newblock Proactive deepfake detection using gan-based visible watermarking.
\newblock {\em ACM Transactions on Multimedia Computing, Communications and Applications}, 2023.

\bibitem{nie2022diffusion}
Weili Nie, Brandon Guo, Yujia Huang, Chaowei Xiao, Arash Vahdat, and Anima Anandkumar.
\newblock Diffusion models for adversarial purification, 2022.

\bibitem{qi2024investigating}
Biqing Qi, Junqi Gao, Yiang Luo, Jianxing Liu, Ligang Wu, and Bowen Zhou.
\newblock Investigating deep watermark security: An adversarial transferability perspective, 2024.

\bibitem{qiao2023novel}
Tong Qiao, Yuyan Ma, Ning Zheng, Hanzhou Wu, Yanli Chen, Ming Xu, and Xiangyang Luo.
\newblock A novel model watermarking for protecting generative adversarial network.
\newblock {\em Computers \& Security}, 127:103102, 2023.

\bibitem{Rombach_2022_CVPR}
Robin Rombach, Andreas Blattmann, Dominik Lorenz, Patrick Esser, and Bj\"orn Ommer.
\newblock High-resolution image synthesis with latent diffusion models.
\newblock In {\em Proceedings of the IEEE/CVF Conference on Computer Vision and Pattern Recognition}, pages 10684--10695, 6 2022.

\bibitem{saberi2024robustness}
Mehrdad Saberi, Vinu~Sankar Sadasivan, Keivan Rezaei, Aounon Kumar, Atoosa Chegini, Wenxiao Wang, and Soheil Feizi.
\newblock Robustness of ai-image detectors: Fundamental limits and practical attacks, 2024.

\bibitem{sharma2023review}
Sunpreet Sharma, Ju~Jia Zou, Gu~Fang, Pancham Shukla, and Weidong Cai.
\newblock A review of image watermarking for identity protection and verification.
\newblock {\em Multimedia Tools and Applications}, pages 1--63, 2023.

\bibitem{soori2023artificial}
Mohsen Soori, Behrooz Arezoo, and Roza Dastres.
\newblock Artificial intelligence, machine learning and deep learning in advanced robotics, a review.
\newblock {\em Cognitive Robotics}, 2023.

\bibitem{2019stegastamp}
Matthew Tancik, Ben Mildenhall, and Ren Ng.
\newblock Stegastamp: Invisible hyperlinks in physical photographs.
\newblock In {\em IEEE Conference on Computer Vision and Pattern Recognition (CVPR)}, 2020.

\bibitem{tao2014robust}
Hai Tao, Li~Chongmin, Jasni~Mohamad Zain, and Ahmed~N Abdalla.
\newblock Robust image watermarking theories and techniques: A review.
\newblock {\em Journal of applied research and technology}, 12(1):122--138, 2014.

\bibitem{touvron2023llama}
Hugo Touvron, Louis Martin, Kevin Stone, Peter Albert, Amjad Almahairi, Yasmine Babaei, Nikolay Bashlykov, Soumya Batra, Prajjwal Bhargava, Shruti Bhosale, Dan Bikel, Lukas Blecher, Cristian~Canton Ferrer, Moya Chen, Guillem Cucurull, David Esiobu, Jude Fernandes, Jeremy Fu, Wenyin Fu, Brian Fuller, Cynthia Gao, Vedanuj Goswami, Naman Goyal, Anthony Hartshorn, Saghar Hosseini, Rui Hou, Hakan Inan, Marcin Kardas, Viktor Kerkez, Madian Khabsa, Isabel Kloumann, Artem Korenev, Punit~Singh Koura, Marie-Anne Lachaux, Thibaut Lavril, Jenya Lee, Diana Liskovich, Yinghai Lu, Yuning Mao, Xavier Martinet, Todor Mihaylov, Pushkar Mishra, Igor Molybog, Yixin Nie, Andrew Poulton, Jeremy Reizenstein, Rashi Rungta, Kalyan Saladi, Alan Schelten, Ruan Silva, Eric~Michael Smith, Ranjan Subramanian, Xiaoqing~Ellen Tan, Binh Tang, Ross Taylor, Adina Williams, Jian~Xiang Kuan, Puxin Xu, Zheng Yan, Iliyan Zarov, Yuchen Zhang, Angela Fan, Melanie Kambadur, Sharan Narang, Aurelien Rodriguez, Robert Stojnic, Sergey Edunov, and Thomas
  Scialom.
\newblock Llama 2: Open foundation and fine-tuned chat models, 2023.

\bibitem{wen2023treering}
Yuxin Wen, John Kirchenbauer, Jonas Geiping, and Tom Goldstein.
\newblock Tree-ring watermarks: Fingerprints for diffusion images that are invisible and robust, 2023.

\bibitem{zhang2019robust}
Kevin~Alex Zhang, Lei Xu, Alfredo Cuesta-Infante, and Kalyan Veeramachaneni.
\newblock Robust invisible video watermarking with attention, 2019.

\bibitem{zhang2024robust}
Lijun Zhang, Xiao Liu, Antoni~Viros Martin, Cindy~Xiong Bearfield, Yuriy Brun, and Hui Guan.
\newblock Robust image watermarking using stable diffusion.
\newblock {\em arXiv preprint arXiv:2401.04247}, 2024.

\bibitem{zhang2023law}
Z.~Zhang.
\newblock Legal challenges in ai-generated content.
\newblock {\em AI \& Law Review}, 12(4):123--130, 2023.

\bibitem{zhao2023invisible}
Xuandong Zhao, Kexun Zhang, Zihao Su, Saastha Vasan, Ilya Grishchenko, Christopher Kruegel, Giovanni Vigna, Yu-Xiang Wang, and Lei Li.
\newblock Invisible image watermarks are provably removable using generative ai, 2023.

\bibitem{zhong2023brief}
Xin Zhong, Arjon Das, Fahad Alrasheedi, and Abdullah Tanvir.
\newblock A brief, in-depth survey of deep learning-based image watermarking.
\newblock {\em Applied Sciences}, 13(21):11852, 2023.

\end{thebibliography}

\end{document}